\documentclass[unnumsec,webpdf,contemporary,large,numbered]{oup-authoring-template}%
\usepackage{booktabs} 
\graphicspath{{Fig/}}
\usepackage{booktabs}
\usepackage{makecell}
\usepackage{multirow}
\usepackage{array}
\theoremstyle{thmstyleone}%
\theoremstyle{thmstyletwo}%
\theoremstyle{thmstylethree}%

\begin{document}

\journaltitle{Journal Title Here}
\DOI{DOI added during production}
\copyrightyear{YEAR}
\pubyear{YEAR}
\vol{XX}
\issue{x}
\access{Published: Date added during production}
\appnotes{Paper}

\firstpage{1}

%\subtitle{Subject Section}

\title[Short Article Title]{Constraint-Aware Optimization for Robust Protein Stability Prediction}

\author[1,$\ast$]{A Shivram\ORCID{0000-0002-9195-2802}}
\author[1]{Aneesh S. Chivukula\ORCID{0000-0002-0445-4435}}
\author[1]{Manik Gupta\ORCID{0000-0002-4977-4299}}
\author[2,$\ast$]{Sourav Chowdhury\ORCID{0000-0002-1148-2995}}

\address{
$^{1}$Department of Computer Science and Information Systems,
$^{2}$Department of Biological Sciences,\\
Birla Institute of Technology and Science Pilani, Hyderabad Campus,
Hyderabad, 500078, Telangana, India
}
\corresp[$\ast$]{Corresponding authors. 
A Shivram, 
\href{mailto:p20230075@hyderabad.bits-pilani.ac.in}
{p20230075@hyderabad.bits-pilani.ac.in}; 
Sourav Chowdhury, 
\href{mailto:sourav.chowdhury@hyderabad.bits-pilani.ac.in }
{sourav.chowdhury@hyderabad.bits-pilani.ac.in }}
\received{Date}{0}{Year}
\revised{Date}{0}{Year}
\accepted{Date}{0}{Year}

%\editor{Associate Editor: Name}

%\abstract{
%\textbf{Motivation:} .\\
%\textbf{Results:} .\\
%\textbf{Availability:} .\\
%\textbf{Contact:} \href{name@email.com}{name@email.com}\\
%\textbf{Supplementary information:} Supplementary data are available at \textit{Journal Name}
%online.}

\abstract{%
\textbf{Motivation:}{} Multimodal $\Delta\Delta G$ predictors integrating protein language models with inverse-folding representations achieve strong in-distribution accuracy on the Megascale dataset but exhibit limited robustness on out-of-distribution (OOD) proteins, persistent forward-reverse bias on paired-mutation benchmarks, and under-representation of rare stabilizing mutations. Existing approaches address these limitations primarily through additional architectural components, leaving optimization-level intervention comparatively underexplored.\\
\textbf{Results:}{} We introduce a constraint-aware optimization framework combining Balanced Mean Squared Error, a Siamese anti-symmetric regularizer, and a novel OOD-margin consistency loss on the per-position feature representation, requiring no architectural changes to the SPURS backbone. Across eleven benchmarks and three random seeds, the framework improves Spearman correlation on S669 from 0.486 to 0.540 ($\sigma=0.002$ across seeds), matching the published SPURS baseline (0.50) without architectural modification, and on S461 from 0.653 to 0.711, with consistent smaller gains on five additional OOD datasets. A controlled diagnostic on Ssym reveals that anti-symmetric training does not eliminate systematic forward-reverse bias, indicating that gains arise through implicit regularization rather than exact thermodynamic constraint enforcement.\\
}

\maketitle

%\begin{epigraph}
%Epigraph text. Ximporem qui reperov idempedit modio. Bisto imagnatem quae aceptis
%nobitae quid eum rae adignis quias-sit vellacc uptatur sunt quis rentis eaquasit alia deliquam
%rec-to consed unt. Empor sum ratur ressimusdae. Nam fugiae.
%\source{Epigraph source}
%\end{epigraph}

\section{Introduction}

Predicting the thermodynamic effects of point mutations remains a central challenge in computational biophysics, molecular engineering, and protein science \cite{horne2022recent, pucci2022artificial,qiu2024review}. Mutation-induced changes in folding free energy are commonly quantified through the changes in Gibbs free energy upon mutation ($\Delta\Delta G$), which provides a direct measure of whether a mutation stabilizes or destabilizes the folded state of protein and therefore directly influences catalytic efficiency, conformational robustness, ligand binding, aggregation propensity, and evolutionary fitness. Accurate estimation of ($\Delta\Delta G$) is critical across a broad range of applications, including therapeutic protein optimization, enzyme engineering, evolutionary analysis, pathogenicity interpretation, and understanding mechanisms of antimicrobial resistance. Despite decades of work, reliably predicting mutation-induced changes remains difficult because protein energetics are governed by highly coupled structural, evolutionary, and thermodynamic interactions that are challenging to capture with simplified statistical or physical models.

Historically, computational stability prediction relied primarily on empirical force fields and physics-based energy functions implemented in frameworks such as FoldX \cite{schymkowitz2005foldx} and Rosetta \cite{leaver2011rosetta3}. Although these approaches provide physically interpretable approximations of molecular energetics, their predictive accuracy is often constrained by limited conformational sampling, approximations in the energy landscape, and substantial computational expenses.  Early machine learning approaches subsequently incorporated combinations of sequence-derived, structural, and physicochemical descriptors to improve predictive performance. Recent advances in protein foundation models have substantially expanded their representational capacity. Large-scale protein language models (pLMs) trained on evolutionary sequence corpora, including ESM-family architectures \cite{lin2023evolutionary}, learn rich contextual representations that implicitly encode evolutionary conservation, coevolutionary coupling, and functional constraints. In parallel, inverse-folding and geometric learning models, such as ProteinMPNN \cite{dauparas2022robust,dieckhaus2024transfer}, incorporate explicit three-dimensional structural information via residue-level spatial representations. Together with the release of increasingly large experimental mutation datasets, these pretrained sequence and structural representations have enabled a new generation of multimodal ($\Delta\Delta G$) predictors that integrate learned evolutionary and geometric priors within unified deep learning frameworks.

Among these datasets, the Megascale cDNA-proteolysis dataset introduced by Tsuboyama \cite{tsuboyama2023mega} represents a particularly important advance. Megascale contains more than 770,000 experimentally measured stability changes across 479 protein domains, providing approximately two orders of magnitude more mutational data than previous curated resources. The availability of this dataset has enabled multimodal deep learning systems such as ThermoMPNN \cite{dieckhaus2024transfer}, ProSTAGE \cite{li2024prostage}, PROSTATA \cite{umerenkov2023prostata}, and SPURS \cite{li2025generalizable} to achieve substantial improvements over earlier approaches. SPURS, in particular, integrates ESM2 \cite{lin2023evolutionary} sequence representations with ProteinMPNN \cite{dauparas2022robust} structural embeddings through a lightweight adapter architecture and currently represents one of the strongest publicly reported baselines across multiple protein stability benchmarks. Most recently, JanusDDG \citep{barducci2026janusddg} reported a sequence-only predictor that combines ESM2 embeddings with a bidirectional cross-attention architecture, enforcing antisymmetry as a hard architectural constraint via a Siamese forward pass and a transitivity loss.

Despite these advances, important limitations remain in the robustness and physical reliability of multimodal stability predictors. Most notably, performance on out-of-distribution (OOD) benchmarks remains substantially lower than in-distribution performance on Megascale. Even SPURS report Spearman correlations of only approximately 0.5 on the widely used S669 benchmark \cite{pancotti2022predicting}, which was specifically constructed to contain proteins absent from the training distribution. This gap suggests that increased architectural complexity alone is insufficient to guarantee robust thermodynamic generalization.

\begin{figure*}[t]
\centering
\includegraphics[width=0.9\textwidth]{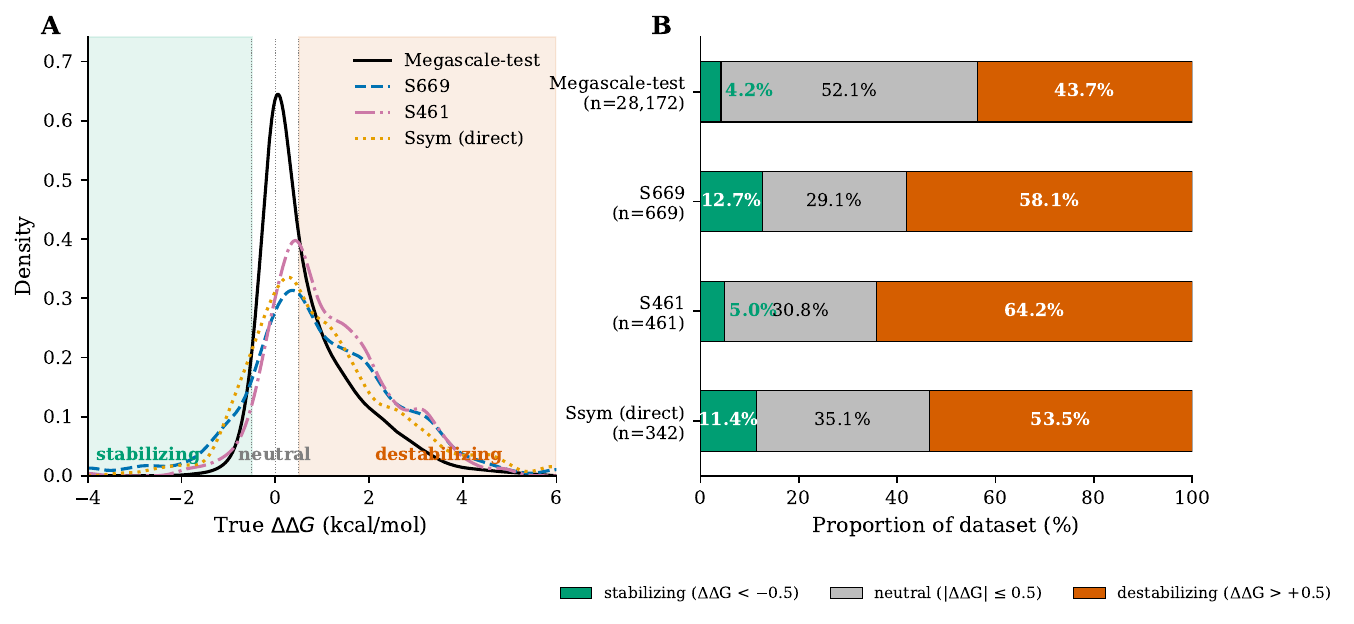}
\caption{
Distributional imbalance across protein stability benchmarks. 
(A) Density distributions of experimental $\Delta\Delta G$ values for representative in-distribution and out-of-distribution benchmarks. 
(B) Fraction of stabilizing, neutral, and destabilizing mutations across datasets. Stabilizing mutations constitute a substantially smaller fraction of the data distribution, motivating the use of imbalance-aware optimization objectives.
}
\label{fig:imbalance}
\end{figure*}
One contributing factor is the strongly imbalanced nature of protein stability datasets. Natural mutational landscapes are heavily skewed towards neutral and destabilizing variants, whereas highly stabilizing mutations are comparatively rare. Across the major training and benchmark datasets, stabilizing substitutions
($\Delta\Delta G < -0.5$~kcal/mol) constitute only $4$-$13\%$ of measured variants, $4.2\%$ on the Megascale training set, $12.7\%$ on S669, $5.0\%$ on S461, and $11.4\%$ on Ssym (direct) (Fig.~\ref{fig:imbalance}), despite representing the mutations of greatest practical value for protein engineering and clinical variant interpretation \cite{sanavia2020limitations}. Under standard regression objectives such as mean squared error or Huber loss, optimization is dominated by the densely populated destabilizing region of the target distribution. Consequently, the optimizer can minimize aggregate regression error through conservative near-zero predictions while assigning relatively limited learning capacity to the rare stabilizing tail. Recent studies have increasingly highlighted the importance of imbalance, mutation-type coverage, and stabilizing-mutation recall in thermodynamic prediction benchmarks \cite{caldararu2021three}, yet most current foundation model-based predictors still optimize regression objectives that remain largely agnostic to the underlying label distribution.

A second limitation arises from the absence of explicit thermodynamic consistency constraints during training. A physically reliable $\Delta\Delta G$ predictor should ideally satisfy the anti-symmetry relationship: 
\begin{equation}
\Delta\Delta G_{\mathrm{wt}\rightarrow\mathrm{mut}}
\;=\;
-\,\Delta\Delta G_{\mathrm{mut}\rightarrow\mathrm{wt}},
\label{eq:antisym}
\end{equation}
because forward and reverse mutations correspond to the same thermodynamic process traversed in opposite directions. This property forms the basis of the Ssym benchmark \cite{pucci2018quantification} and has motivated earlier antisymmetric neural architectures developed prior to the current foundation model era \cite{benevenuta2021antisymmetric}. Although several earlier architectures incorporated thermodynamic reversibility through architectural coupling and data augmentation strategies, the role of optimization-level thermodynamic regularization in modern multimodal foundation model-based stability predictors remains comparatively underexplored. As a result, forward and reverse predictions may exhibit substantial directional bias despite preserving overall rank correlation. In our measurements on Ssym, forward-reverse predictions consistently display systematic offsets of approximately 0.3-0.6 kcal/mol across random seeds.

A third, broader limitation is that the standard training objective provides no explicit defense against representation drift on OOD proteins. The multimodal feature vectors produced by ESM-2 and ProteinMPNN are themselves distribution-shifted between Megascale-train and any OOD benchmark: even before the MLP head, the encoder outputs on S669 proteins are statistically different from those on Megascale train. A model trained without any explicit defense against small feature perturbations may overfit in-distribution feature statistics, generalizing poorly when the same encoder produces slightly different feature distributions on unseen proteins. Despite increasing attention to OOD robustness in machine learning broadly \cite{krueger2021out,sagawa2019distributionally}, this specific failure mode has not been systematically addressed in foundation model-based $\Delta\Delta G$ prediction pipelines.
\begin{figure*}[hh]
\centering
\includegraphics[width=\textwidth]{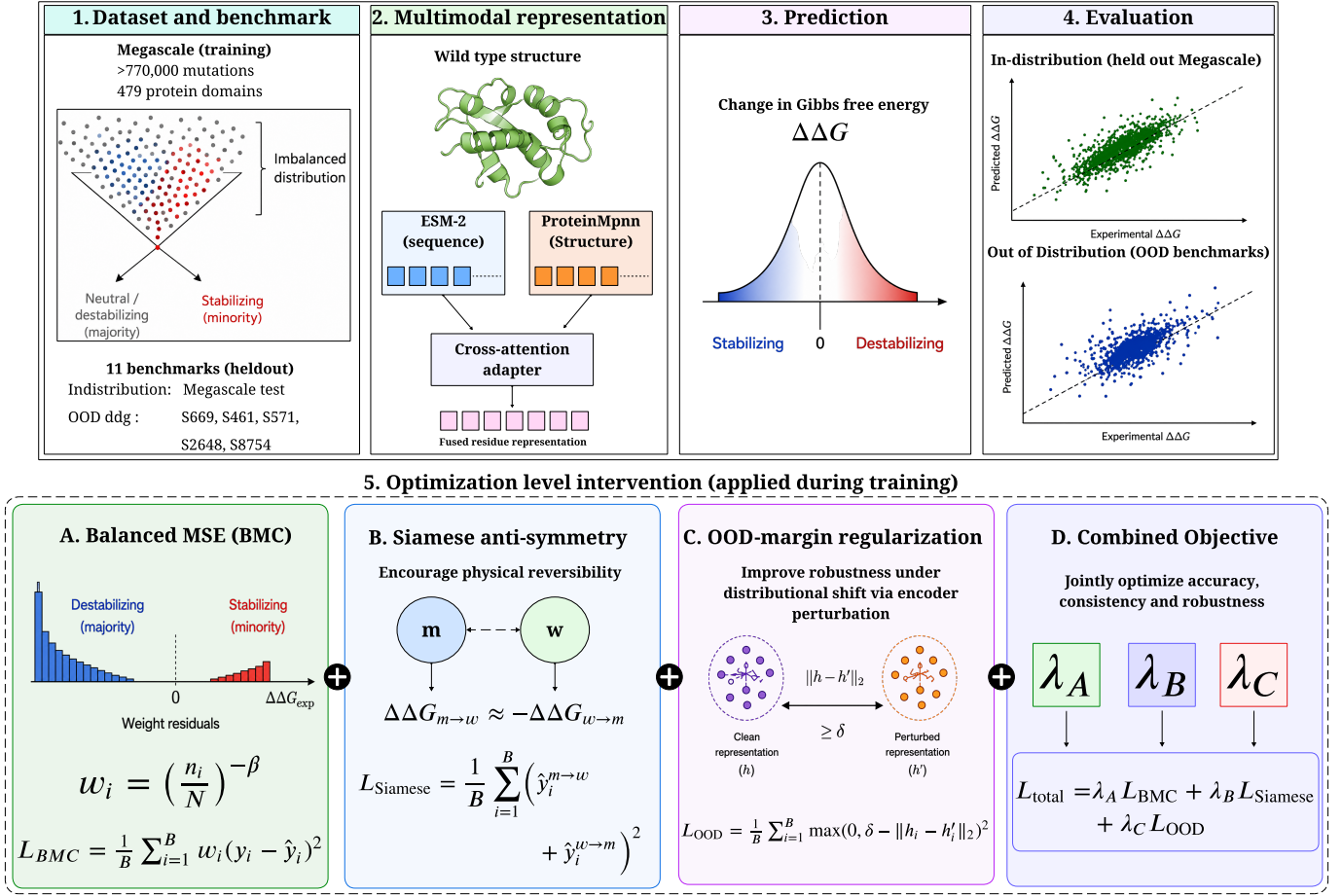}
\caption{
Overview of the proposed constrained optimization framework for multimodal protein stability prediction. Sequence embeddings from ESM-2 and structure-aware representations from ProteinMPNN are integrated through cross-attention fusion to predict mutation-induced changes in Gibbs free energy ($\Delta\Delta G$). The framework is trained on Megascale and evaluated on both in-distribution and out-of-distribution (OOD) benchmarks. During training, Balanced Mean Squared Error (BMC), Siamese anti-symmetric regularization, and OOD-margin consistency regularization are jointly applied to improve robustness and thermodynamic generalization.
}
\label{fig:workflow}
\end{figure*}

In this work, we investigate whether these limitations can be addressed through optimization-level interventions layered on top of an existing multimodal foundation model backbone without modifying the underlying architecture. Our motivation is both practical and scientific: retraining large multimodal protein models from scratch is computationally expensive, whereas loss-level interventions can be integrated into existing systems with minimal architectural overhead. We therefore focus specifically on restructuring the optimization landscape rather than introducing additional fusion modules or deeper attention mechanisms.

We study three complementary loss-level interventions. First, we adapt the Balanced Mean Squared Error (BMC) \cite{ren2022balanced} to protein stability prediction. BMC treats regression targets as samples from a continuous distribution and dynamically reweights gradients to increase optimization pressure on underrepresented regions of label space, particularly highly stabilizing mutations. Second, we introduce a Siamese anti-symmetric objective in which both forward and reverse mutations are evaluated via shared-weight forward passes, and their summed predictions are explicitly penalized during training. Third, we propose an \emph{OOD-margin} loss specifically designed for a foundation model-based stability predictor: small Gaussian perturbations are applied to the encoder's per-position feature representation, and the squared difference between the clean and perturbed prediction is penalized. The OOD-margin loss is a first-order consistency regularizer that encourages the MLP head to produce predictions that are stable under small representation drift, directly addressing the third limitation identified above. Together, these interventions aim to improve robustness to imbalanced thermodynamic distributions, introduce lightweight thermodynamic regularization, and defend against in-distribution feature-level overfitting.

Across three random seeds and eleven public benchmarks, the proposed optimization framework consistently improves performance on challenging out-of-distribution stability benchmarks. In particular, the combined objective increases Spearman correlation on S669 from $0.486$ to $0.540$ and on S461 from $0.653$ to $0.711$, while maintaining competitive performance across additional OOD datasets. These gains are accompanied by a modest reduction in in-distribution Megascale performance, consistent with a redistribution of optimization capacity toward structurally challenging OOD regimes. Interestingly, diagnostic analysis on Ssym reveals that anti-symmetric training modifies but does not fully eliminate systematic forward-reverse bias, suggesting that the primary benefit of the proposed framework arises from improved optimization dynamics and implicit regularization rather than exact enforcement of thermodynamic consistency. We further report several carefully characterized negative results, including unsuccessful attempts with auxiliary $K_{50}$ supervision, structural relaxation features, and explicit batch-level bias correction, to better delineate which forms of physical supervision meaningfully contribute to robust thermodynamic generalization.

Taken together, our results provide a systematic analysis of how distribution-aware, reversibility-aware, and representation-stability-aware optimization influence modern multimodal protein stability predictors. More broadly, the findings suggest that physically motivated loss design can improve out-of-distribution robustness and predictive reliability even when exact thermodynamic consistency remains unresolved. An overview of the proposed multimodal optimization framework is shown in Fig.~\ref{fig:workflow}
\section{Methods}
\label{sec:Methods}
\subsection{Datasets}
\label{sec:datasets}
Models were trained on the Megascale cDNA-proteolysis dataset \cite{tsuboyama2023mega}, which contains experimentally measured $\Delta\Delta G$ values for mutations across $479$ protein domains. Following prior work, we restricted training to single-point mutations and applied MMseqs2 filtering at $30\%$ sequence identity to remove near-duplicate proteins. After filtering, the training, validation, and held-out test splits contained $212$, $29$, and $28$ proteins, respectively, corresponding to approximately $190{,}000$ mutations in the training split. 

For out-of-distribution evaluation, we used ten external benchmarks spanning both thermodynamic stability ($\Delta\Delta G$) and thermal stability ($\Delta T_{\mathrm{m}}$) measurements, including S669
\cite{pancotti2022predicting}, S461, Ssym \cite{pucci2018quantification,benevenuta2021antisymmetric}, S571, S783, FireProt-HF \cite{stourac2021fireprotdb}, S2648, S4346, and S8754. Benchmark proteins were nonredundant with the Megascale training split after MMseqs2 filtering.

\subsection{Backbone architecture}
\label{sec:backbone}

Our implementation is inspired by the SPURS framework \cite{lin2023evolutionary}, which combines sequence representations from ESM-2 (\texttt{esm2\_t33\_650M\_UR50D}) \cite{lin2023evolutionary} with structural features from ProteinMPNN with the published \texttt{v\_48\_020}
weights ($k=48$ nearest neighbors, hidden dimension $128$) \cite{dauparas2022robust} through a lightweight
residual adapter. Following the overall SPURS design philosophy, we used ESM-2 as the sequence encoder and ProteinMPNN as the structural encoder, with ProteinMPNN features injected into the upper layers of ESM-2 via residual fusion. 

For each mutation, the fused per-residue representation at the mutated position was passed through a three-layer multilayer perceptron with GELU activations and dropout to produce amino-acid-specific scores. The predicted $\Delta\Delta G$ for a mutation 
$\mathrm{wt}\rightarrow\mathrm{mut}$ at position $p$ was computed as

\begin{equation}
\hat{\Delta\Delta G}_{\mathrm{wt}\rightarrow\mathrm{mut}}
=
\mathrm{MLP}(h_p)_{\mathrm{mut}}
-
\mathrm{MLP}(h_p)_{\mathrm{wt}},
\label{eq:ddg_gather}
\end{equation}
where $h_p$ denotes the fused residue representation at the mutated position. This gather-based formulation additionally enables efficient evaluation of reverse mutations required for the Siamese anti-symmetric objective.
\subsection{Loss objectives}
\label{sec:loss}

\paragraph{Balanced MSE (BMC).}
The Megascale training distribution is heavily imbalanced toward neutral and destabilizing mutations: $52.1\%$ of training mutations fall within the neutral band ($|\Delta\Delta G| \leq 0.5$~kcal/mol), $43.7\%$ are destabilizing ($\Delta\Delta G> +0.5$~kcal/mol), and only $4.2\%$ are stabilizing ($\Delta\Delta G< -0.5$~kcal/mol). Under a standard regression objective, the optimizer can minimize aggregate loss by predicting toward the dense neutral median, leaving the rare stabilizing tail systematically under-represented in gradients. To reduce optimization bias toward these dominant regions of label space, we replaced the standard regression objective with Balanced Mean Squared Error (BMC) \cite{ren2022balanced}. BMC formulates regression as a distributional-aware contrastive objective:

\begin{equation}
\mathcal{L}_{\mathrm{BMC}}
=
-\frac{1}{B}\sum_{i=1}^{B}
\log
\frac{
\exp\!\left(
-\frac{(\hat{y}_i-y_i)^2}
{2\sigma_{\mathrm{BMC}}^2}
\right)
}{
\sum_{j=1}^{B}
\exp\!\left(
-\frac{(\hat{y}_i-y_j)^2}
{2\sigma_{\mathrm{BMC}}^2}
\right)
},
\label{eq:bmc}
\end{equation}

where $\hat{y}_i$ and $y_i$ denote predicted and experimental $\Delta\Delta G$ values, and $\sigma_{\mathrm{BMC}}$ is a learnable noise-scale parameter. This formulation dynamically increases optimization emphasis on the sparsely represented region of the label distribution, particularly highly stabilizing mutations. 

\paragraph{Siamese anti-symmetric regularization.}
To encourage thermodynamic consistency, we introduced a Siamese anti-symmetric objective in which both forward ($\mathrm{wt}\rightarrow\mathrm{mut}$) and reverse ($\mathrm{mut}\rightarrow\mathrm{wt}$) mutations were evaluated through shared network weights. Reverse mutations were generated by swapping the wild-type and mutant residue identities at the mutated position. The anti-symmetric loss was defined as 
\begin{equation}
\mathcal{L}_{\mathrm{sym}}
=
\frac{1}{B}\sum_{i=1}^{B}
\left(
\hat{\Delta\Delta G}^{(i)}_{\mathrm{wt}\rightarrow\mathrm{mut}}
+
\hat{\Delta\Delta G}^{(i)}_{\mathrm{mut}\rightarrow\mathrm{wt}}
\right)^2,
\label{eq:sym_loss}
\end{equation}

where $B$ denotes the batch size. This objective was applied as a soft regularizer throughout training with
$\lambda_{\mathrm{sym}} = 0.5$. The diagnostic results in Section~\ref{sec:res_antisym} characterize what this loss actually achieves on Ssym.

\paragraph{OOD-margin regularization}
To improve robustness against representation drift, we introduce an OOD-margin consistency objective based on perturbations of the fused per-residue representation. For each training example, Gaussian noise $\boldsymbol{\delta}_i \sim
\mathcal{N}(0,\sigma_{\mathrm{OOD}}^2\mathbf{I})$ was added to the per-position feature representation following the encoder forward pass, and the perturbed representation was re-evaluated through the prediction head to obtain a second prediction $\tilde{y}_i$. The OOD-margin loss was defined as

\begin{equation}
\mathcal{L}_{\mathrm{OOD}}
=
\frac{1}{B}\sum_{i=1}^{B}
\left(
\hat{y}_i-\tilde{y}_i
\right)^2,
\label{eq:ood_margin}
\end{equation}

where $\hat{y}_i$ and $\tilde{y}_i$ denote predictions from the clean and perturbed representations, respectively.The encoder pass is not repeated, so the additional cost of this loss is only the MLP head forward pass and a small noise sample, increasing per-step training time by approximately $10\%$. The hyperparameter $\sigma_{\mathrm{OOD}}$ controls the radius of the consistency neighborhood around each per-position representation. We tune
$\sigma_{\mathrm{OOD}} \in \{0.10, 0.20, 0.50\}$ (Section~\ref{sec:res_sensitivity}) and select $\sigma_{\mathrm{OOD}} = 0.20$ for the headline configuration~\textbf{E}, weighted by $\lambda_{\mathrm{OOD}} = 0.5$. The loss penalizes high sensitivity of the prediction to small perturbations of the feature representation, which acts as a first-order regularizer against feature-statistics overfitting.

\paragraph{Total training objective.}
The total objective for configuration~\textbf{E} is
\begin{equation}
\mathcal{L}_{\mathrm{total}}
\;=\;
\mathcal{L}_{\mathrm{BMC}}
\;+\;
\lambda_{\mathrm{sym}}\,\mathcal{L}_{\mathrm{sym}}
\;+\;
\lambda_{\mathrm{OOD}}\,\mathcal{L}_{\mathrm{OOD}},
\label{eq:total}
\end{equation}
with $\lambda_{\mathrm{sym}} = \lambda_{\mathrm{OOD}} = 0.5$ throughout.
Configurations~\textbf{A}--\textbf{D} use subsets of this objective
as specified in Section~\ref{sec:results}.

\subsection{Training and evaluation}
Models were trained using AdamW \cite{loshchilov2018decoupled} with learning rate $10^{-4}$ and early stopping based on validation Spearman correlation. All the primary experiments were performed across three random seeds. 

Performance was evaluated using Spearman correlation, Pearson correlation, root-mean-square error, and mean absolute error between predicted and experimental
$\Delta\Delta G$ values. Following prior protein stability benchmarks \cite{pancotti2022predicting}, Spearman correlation was used as the primary metric. Results are reported as mean $\pm$ standard deviation across three runs.
Models were implemented in PyTorch. Code, checkpoints, and processed benchmark splits will be released upon publication.
\section{Results and discussion}
\label{sec:results}
To systematically evaluate the impact of optimization-level intervention on multimodal stability prediction, we train five configurations of the backbone (ESM-2 integrated with ProteinMPNN) inspired by SPURS on a deduplicated subset of the Megascale dataset from scratch. The configurations differ exclusively in their objective functions: \textbf{A}, the baseline Huber loss reproducing the original architecture; \textbf{B}, Balanced Mean Squared Error (BMC) alone; \textbf{C}, Huber loss augmented with a Siamese regularizer penalizing forward-reverse inconsistency; \textbf{D}, the combined BMC and Siamese objective; and \textbf{E}, the full three-component objective combining BMC, Siamese training, and the OOD-margin loss (input-noise consistency on the per-position feature representation). Each configuration is evaluated across three independent random initializations (seeds $42$, $43$, $44$), with performance reported on the held-out Megascale test split and 10 independent OOD benchmarks. 

\subsection{Robustness and Out-of-Distribution Generalization}
\label{sec:res_main}
The central empirical finding of this study is that restructuring the optimization landscape yields large, reproducible gains in OOD generalization. As detailed in Table~\ref{tab:main}, the combined three-component configurations~\textbf{E} achieve a Spearman correlation $0.540 \pm 0.002$ on the challenging S669 benchmark, an improvement of $+0.054$ over the SPURS baseline configuration. A parallel improvement of $+0.058$ is observed on S461 ($0.653 \to 0.711$), with consistent, smaller magnitude gains S4346 ($+0.010$), S8754 ($+0.009$), and Ssym-direct ($+0.010$). The seed-level standard deviation of the S669 gain is $0.002$, an order of magnitude smaller than the gain itself, providing strong evidence that the improvement is not an artifact of seed selection.

\begin{table*}[t]
\centering
\small
\setlength{\tabcolsep}{6pt}
\renewcommand{\arraystretch}{1.1}
\caption{Spearman rank correlation between predicted and experimental $\Delta\Delta G$ across eleven benchmarks for five optimization configurations. All values are reported as mean~$\pm$~standard deviation across three independent random seeds, with the best-performing configuration for each benchmark shown in bold. Configuration~\textbf{A} reproduces the original training objective, configurations~\textbf{B} to \textbf{D} apply Balanced MSE and Siamese anti-symmetric losses individually and in combination; configuration~\textbf{E} adds the OOD-margin input-noise consistency regularizer at $\sigma=0.20$. $n$ denotes the number of mutations in each benchmark.}
\label{tab:main}
\resizebox{0.95\textwidth}{!}{%
\begin{tabular}{lrccccc}
\toprule
Benchmark & $n$ & A: baseline & B: + BMC & C: + Siam. & D: + BMC + Siam. & E: full \\
\midrule
\multicolumn{7}{l}{\textit{Out-of-distribution (primary)}}\\
\quad S669             &   669 & $0.486 \pm 0.012$ & $0.505 \pm 0.013$ & $0.502 \pm 0.008$ & $0.517 \pm 0.007$ & $\mathbf{0.540 \pm 0.002}$ \\
\quad S461             &   461 & $0.653 \pm 0.010$ & $0.680 \pm 0.005$ & $0.671 \pm 0.012$ & $0.683 \pm 0.005$ & $\mathbf{0.711 \pm 0.006}$ \\
\quad S571 ($\Delta T_{\mathrm{m}}$) & 571 & $0.447 \pm 0.006$ & $\mathbf{0.466 \pm 0.005}$ & $0.439 \pm 0.009$ & $0.459 \pm 0.005$ & $0.465 \pm 0.009$ \\
\midrule
\multicolumn{7}{l}{\textit{Out-of-distribution (auxiliary)}}\\
\quad Ssym (direct)    &   342 & $0.710 \pm 0.007$ & $0.720 \pm 0.004$ & $0.703 \pm 0.005$ & $0.716 \pm 0.011$ & $\mathbf{0.726 \pm 0.024}$ \\
\quad Ssym (inverse)   &   342 & $0.609 \pm 0.020$ & $\mathbf{0.619 \pm 0.005}$ & $0.598 \pm 0.018$ & $0.608 \pm 0.010$ & $0.608 \pm 0.017$ \\
\quad FireProt-HF      &  2578 & $0.644 \pm 0.005$ & $\mathbf{0.652 \pm 0.005}$ & $0.643 \pm 0.002$ & $0.649 \pm 0.002$ & $0.650 \pm 0.005$ \\
\quad S783             &   783 & $0.696 \pm 0.009$ & $\mathbf{0.701 \pm 0.007}$ & $0.699 \pm 0.001$ & $0.695 \pm 0.008$ & $0.697 \pm 0.004$ \\
\quad S8754            &  8228 & $0.605 \pm 0.007$ & $0.621 \pm 0.002$ & $0.612 \pm 0.010$ & $0.619 \pm 0.003$ & $\mathbf{0.628 \pm 0.002}$ \\
\quad S2648            &  2648 & $0.666 \pm 0.009$ & $\mathbf{0.680 \pm 0.005}$ & $0.669 \pm 0.008$ & $0.679 \pm 0.001$ & $\mathbf{0.682 \pm 0.004}$ \\
\quad S4346            &  3969 & $0.612 \pm 0.007$ & $0.611 \pm 0.002$ & $0.615 \pm 0.005$ & $0.608 \pm 0.002$ & $\mathbf{0.618 \pm 0.001}$ \\
\midrule
\multicolumn{7}{l}{\textit{In-distribution}}\\
\quad Megascale-test   & 28172 & $\mathbf{0.749 \pm 0.004}$ & $0.729 \pm 0.003$ & $\mathbf{0.749 \pm 0.001}$ & $0.727 \pm 0.003$ & $0.713 \pm 0.003$ \\
\bottomrule
\end{tabular}
}
\end{table*}

While these absolute Spearman gains may appear numerically modest, they represent a substantial methodological advance when contextualized against the experimental noise floor. Because ground-truth $\Delta\Delta G$ measurements contain inherent experimental variance across laboratories ($\sim 0.1$--$0.5$~kcal/mol per measurement \cite{montanucci2019natural}, previous theoretical analyses establish that the maximum achievable correlation for any structure-based predictor is bounded at approximately 0.71 \cite{montanucci2019natural}. Consequently, predictive performance must be evaluated relative to this theoretical ceiling rather than a perfect correlation of 1.0. By advancing S669 performance from the $0.486$ baseline to $0.540$, our framework resolves approximately $24\%$ of the remaining addressable performance gap on this benchmark, roughly one quarter
of the headroom left between the SPURS baseline and the
data-noise-imposed ceiling. This indicates that restructuring the optimization landscape enables a highly efficient extraction of the remaining, non-stochastic structural signal that architectural
modifications alone have so far failed to capture. 
\begin{table}[t]
\centering
\resizebox{\columnwidth}{!}{%
\begin{tabular}{lcccccc}
\toprule
Dataset & ESM & ProteinMPNN & ThermoMPNN & Stability Oracle & SPURS & Ours \\
\midrule
S669  & 0.37 & 0.43 & 0.44 & 0.50 & 0.50 & \textbf{0.540$\pm$0.002} \\
S461  & 0.29 & 0.50 & 0.61 & --   & 0.66 & \textbf{0.711$\pm$0.006} \\
S783  & 0.30 & 0.52 & 0.70 & --   & \textbf{0.71} & 0.697$\pm$0.004 \\
S2648 & 0.18 & 0.31 & 0.65 & --   & \textbf{0.69} & 0.682$\pm$0.004 \\
S8754 & 0.16 & 0.24 & 0.61 & --   & \textbf{0.64} & 0.628$\pm$0.002 \\
\bottomrule
\end{tabular}
}
\caption{Spearman rank correlation ($\rho$) between predicted and experimental
$\Delta\Delta G$ for our framework and previously reported predictors on five
out-of-distribution benchmarks. Baseline values for ESM, ProteinMPNN,
ThermoMPNN, Stability Oracle, and SPURS are taken from the respective
publications~\cite{li2025generalizable,dieckhaus2024transfer, diaz2024stability}; dashes denote values not reported in the original
source. Our values are mean $\pm$ s.d. over three seeds for the full
three-component objective (BMC + Siamese + OOD-margin). Best value per row in
\textbf{bold}.}
\label{tab:baseline-comparison}
\end{table}
These OOD improvements coincide with a $0.014$ Spearman reduction on the in-distribution Megascale test split ($0.749 \to 0.713$). This reduction is not an arbitrary failure, but an expected biophysical tradeoff. Standard regression aggressively minimizes error in the densely populated neural regions of the training distribution, thereby maximizing in-distribution metrics at the cost of structural generalization. The proposed loss modifications intentionally perturb this in-distribution memorization, shifting optimization capacity towards transferable structural features required for OOD generalization. From a protein-engineering perspective, the tradeoff favors the use case that actually matters: a model deployed for engineering or variant prioritization will routinely encounter proteins not seen during training, and its OOD reliability is the primary practical concern.

\subsection{Component Ablation: The Additivity of Optimization Interventions}
\label{sec:res_ablation}
To elucidate the distinct mechanisms driving these improvements, we ablate the individual contributions of BMC, the Siamese objective and the OOD-margin loss (Table~\ref{tab:delta}). The results show three distinct regimes of behavior across the OOD benchmarks.

\begin{table}[b]
\centering
\footnotesize
\setlength{\tabcolsep}{4pt}
\renewcommand{\arraystretch}{1.05}
\caption{Mean Spearman improvement relative to the baseline configuration~\textbf{A} across OOD benchmarks where at least one optimization configuration achieves a gain of $\geq +0.01$. Values are averaged across three independent random seeds. The three loss components
contribute roughly additively on the structurally hardest benchmarks (S669, S461), partially overlap on intermediate benchmarks, and contribute non-monotonically on $\Delta T_{\mathrm{m}}$-based benchmarks.}
\label{tab:delta}
\begin{tabular}{lrrrrr}
\toprule
Benchmark & A (baseline) & $\Delta$ B (BMC) & $\Delta$ C (Siam.) & $\Delta$ D (B+C) & \shortstack[c]{$\Delta$ E\\(+OOD)} \\
\midrule
S669             & 0.486 & $+0.019$ & $+0.015$ & $+0.031$ & $+0.054$ \\
S461             & 0.653 & $+0.027$ & $+0.018$ & $+0.030$ & $+0.058$ \\
S571             & 0.447 & $+0.018$ & $-0.008$ & $+0.012$ & $+0.018$ \\
S2648            & 0.666 & $+0.014$ & $+0.003$ & $+0.012$ & $+0.016$ \\
S8754            & 0.605 & $+0.016$ & $+0.007$ & $+0.014$ & $+0.023$ \\
S4346            & 0.612 & $-0.002$ & $+0.003$ & $-0.004$ & $+0.006$ \\
Ssym (direct)    & 0.710 & $+0.010$ & $-0.007$ & $+0.005$ & $+0.015$ \\
\bottomrule
\end{tabular}
\end{table}

On the most structurally challenging $\Delta\Delta G$ benchmarks (S669 and S461), the three contribute roughly additively. On S669, BMC independently contributes $+0.019$ Spearman, while the Siamese regularizer independently contributes $+0.015$, and adding the OOD-margin loss to the BMC+Siamese baseline contributes an additional $+0.023$ on top of D's $+0.031$. The total gain of $+0.054$ is within statistical agreement of the simple sum of independent contributions, suggesting that the three objectives constrain orthogonal aspects of the prediction landscape: BMC reweights the gradients distribution to prevent collapse toward the median, the Siamese objective regularizes the magnitude of prediction when forward and reverse predictions conflict, and the OOD-margin loss penalizes high sensitivity of the output to small perturbations of the per-position feature representation. These mechanisms compose because they act on different aspects of the optimization landscape.

On intermediate benchmarks (S2648, S8754), the OOD-margin loss contributes a smaller additional gain over the BMC+Siamese baseline, roughly $+0.004$ and $+0.009$ respectively. On S571 (a $\Delta T_{\mathrm{m}}$ benchmark), the Siamese loss actively degrades performance ($-0.008$), and the combined configuration's $+0.018$ gain comes essentially from BMC and the OOD-margin loss together rather than from anti-symmetric training. Because the thermodynamic mapping between $\Delta\Delta G$ and $\Delta T_{m}$ is fold-dependent and linear only over narrow regimes, imposing strict magnitude regularization via the Siamese loss inappropriately constrains $\Delta T_{m}$ predictions. This divergence highlights a critical limitation: magnitude-based Siamese regularizers are well-suited to free-energy landscapes but should not be broadly applied to proxy thermostability metrics.

\subsection{Mechanistic Analysis of Thermodynamic Anti-Symmetry}
\label{sec:res_antisym}
A primary motivation for the Siamese objective is the enforcement of physical reversibility. A thermodynamically consistent model must satisfy the relationship: 
\begin{equation}
    \Delta\Delta G_{\mathrm{wt}\rightarrow\mathrm{mut}} = -\Delta\Delta G_{\mathrm{mut}\rightarrow\mathrm{wt}}
\end{equation}

\begin{figure*}[t]
  \centering
  \includegraphics[width=\textwidth]{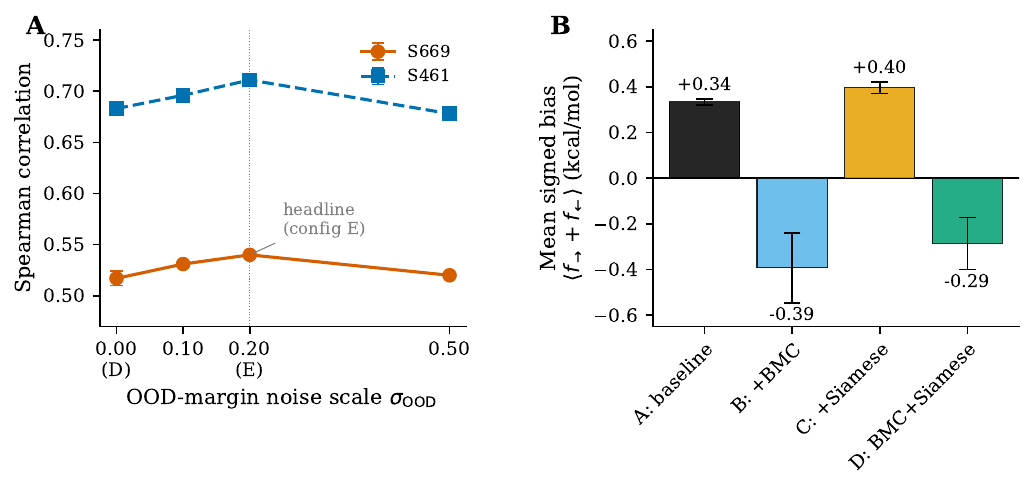}
  \caption{(\textbf{A}) OOD-margin hyperparameter sweep on S669 and S461.
  Error bars on 3-seed configurations (D at $\sigma=0$ and E at
  $\sigma=0.20$); $\sigma=0.10$ and $\sigma=0.50$ are seed-42 only.
  The inverted-U profile shows that the OOD-margin gain is sharply
  localized at $\sigma=0.20$ rather than a generic regularization
  effect. (\textbf{B}) Systematic forward-reverse bias
  $\langle f_{\rightarrow} + f_{\leftarrow}\rangle$ on Ssym, decomposed
  per configuration. Baseline and siamese configurations have positive
  bias; BMC-based configurations flip the sign. The Siamese loss does
  not eliminate the bias.}
  \label{fig:fig1}
\end{figure*}

However, evaluating the forward-reverse sum ($\bar{\varepsilon}_{\mathrm{sym}}$) on the Ssym benchmark reveals a non-obvious failure mode in standard foundational predictors (Figure~\ref{fig:fig1}).

% \begin{table}[h]
% \centering
% \caption{Anti-symmetry diagnostic on the Ssym benchmark. The forward-reverse consistency error (Eq.~\ref{eq:eps_sym}) is decomposed into a systematic directional bias and a residual standard deviation component. Although configurations~\textbf{B} and \textbf{D} exhibit larger raw anti-symmetry violations than the baseline, these increases arise primarily from shifts in systematic bias rather than from degradation of the underlying forward-reverse correlation structure. Pearson correlation between forward predictions and negated reverse predictions is additionally reported; values near $+1.0$ indicate preservation of the anti-symmetric relationship up to
% an additive offset. All quantities are reported as
% mean~$\pm$~standard deviation across three independent random seeds.}
% \label{tab:antisym}
% \label{tab:antisym}
% \begin{tabular}{lrrrr}
% \toprule
% Config & $\bar{\varepsilon}_{\mathrm{sym}}$ & Mean bias & Residual $\sigma$ & Pearson($f_{\to}$, $-f_{\leftarrow}$) \\
% \midrule
% A: baseline    & $0.364 \pm 0.011$ & $+0.335$ & $0.384$ & $+0.882$ \\
% B: + BMC       & $0.622 \pm 0.111$ & $-0.393$ & $0.644$ & $+0.872$ \\
% C: + Siamese   & $0.417 \pm 0.024$ & $+0.396$ & $0.396$ & $+0.884$ \\
% D: combined    & $0.549 \pm 0.077$ & $-0.286$ & $0.626$ & $+0.880$ \\
% \bottomrule
% \end{tabular}
% \end{table}

The baseline model (\textbf{A}) exhibits a systematic directional bias, overestimating stability changes in the forward direction by an average of +0.335 kcal/mol. Crucially, the addition of the Siamese regularizer (\textbf{C}) does not eliminate this bias; it remains at $+0.396$~kcal/mol. The Siamese objective penalizes the squared sum $(f_{\to} + f_{\leftarrow})^2$. When predictions conflict, the resulting gradient uniformly pulls both forward and inverse predictions toward smaller absolute magnitudes rather than forcing them toward a true zero-sum relationship: a model that achieves $f_{\to} = f_{\leftarrow} = +0.2$ kcal/mol has a small Siamese loss but is catastrophically wrong about the physical direction of every mutation effect.

Consequently, the $+0.054$ OOD generalization gain observed on S669 in configuration~\textbf{E} cannot be attributed to the strict enforcement of thermodynamic reversibility, as the network remains physically inconsistent in the same systematic way as the baseline. Instead, we propose using Siamese objective functions as a magnitude regularizer to prevent the BMC and OOD-margin objectives from causing optimization instability when they push predictions towards extreme values. This interpretation is consistent with our observation that the Pearson correlation between $f_{\to}$ and $-f_{\leftarrow}$ is roughly constant ($\sim 0.88$) across all four configurations, even as the absolute violation $\bar{\varepsilon}_{\mathrm{sym}}$ changes substantially. The forward-reverse \emph{relationship} is preserved across configurations; only the additive bias changes. This finding suggests that future loss designs should explicitly target signed-bias correction rather than relying solely on symmetric magnitude penalties to better enforce biophysical consistency. We investigate this direction through the BCAS auxiliary loss reported in Section~\ref{sec:negative}.

We note that concurrent work by Barducci et al.~\citep{barducci2026janusddg} demonstrates that perfect antisymmetry can be enforced \emph{architecturally} by averaging direct and inverse forward passes. This achieves 
PCC$_{d\text{-}r} = -1.00$ and zero bias by construction. Our 
finding is complementary: when antisymmetry is imposed as a soft loss-level regularizer rather than as a hard architectural 
constraint, systematic bias persists at $0.3$-$0.4$~kcal/mol, and our BCAS experiment further shows that explicit bias correction does not improve OOD Spearman. This indicates that loss-level antisymmetric training functions as implicit regularization rather than as a physical constraint enforcement, an interpretation that may not transfer to architecturally constrained models such as JanusDDG \cite{barducci2026janusddg}.

\subsection{Hyperparameter sensitivity of the OOD-margin loss}
\label{sec:res_sensitivity}

The OOD-margin loss is parameterized by the Gaussian noise scale $\sigma$ applied to the per-position feature representation. To characterize the sensitivity of the result to this hyperparameter, we sweep $\sigma \in \{0.10, 0.20, 0.50\}$ at fixed weight, with all other hyperparameters held constant at the configuration~\textbf{D} defaults Figure~\ref{fig:fig1}).

% \begin{table}[h]
% \centering
% \caption{Hyperparameter sensitivity of the OOD-margin loss. Each row
% adds the OOD-margin term to the BMC + Siamese baseline (configuration
% \textbf{D}) with the indicated noise scale $\sigma$ on the per-position
% feature representation. Values at $\sigma=0.10$ and $\sigma=0.50$ are
% seed-42 only; values at $\sigma=0.20$ are mean $\pm$ standard deviation
% across three seeds (the configuration we report as \textbf{E}).
% Intermediate noise scales improve OOD generalization; sufficiently
% large $\sigma$ disrupts optimization. The sweet spot at $\sigma=0.20$
% is sharply localized.}
% \label{tab:sigma_sweep}
% \begin{tabular}{lcccc}
% \toprule
% $\sigma$ & S669 & S461 & Ssym (inverse) & Megascale-test \\
% \midrule
% $0.00$ (\textbf{D})  & $0.517 \pm 0.007$ & $0.683 \pm 0.005$ & $0.608 \pm 0.010$ & $0.727 \pm 0.003$ \\
% $0.10$               & $0.531$           & $0.696$           & $0.620$           & $0.719$           \\
% $\mathbf{0.20}$ (\textbf{E})  & $\mathbf{0.540 \pm 0.002}$ & $\mathbf{0.711 \pm 0.006}$ & $0.608 \pm 0.017$ & $0.713 \pm 0.003$ \\
% $0.50$               & $0.520$           & $0.678$           & $0.606$           & $0.701$           \\
% \bottomrule
% \end{tabular}
% \end{table}
The sweep suggests a localized optimum near $\sigma=0.20$. Increasing $\sigma$ from $0$ (configuration \textbf{D}) to $0.10$ already improves S669 by $+0.014$, and increasing further to $\sigma=0.20$ yields an additional $+0.009$, for a total gain of $+0.023$ over \textbf{D}. However, increasing $\sigma$ to $0.50$ substantially reduces this improvement and further degrades Megascale-test performance. These results suggest that small perturbations encourage the model to rely less strongly on brittle in-distribution feature statistics, thereby improving OOD generalization. In contrast, excessively large perturbations may destabilize the consistency target and reduce the utility of the fine-grained per-position representations. Overall, the non-monotonic behavior indicated that the OOD-margin objective operates most effectively within a limited-noise regime rather than acting as a generic regularizer.

\subsection{Engineering Utility: Stabilizing Mutation Recall}
\label{sec:res_stab}
\begin{figure*}[t]
  \centering
  \includegraphics[width=\textwidth]{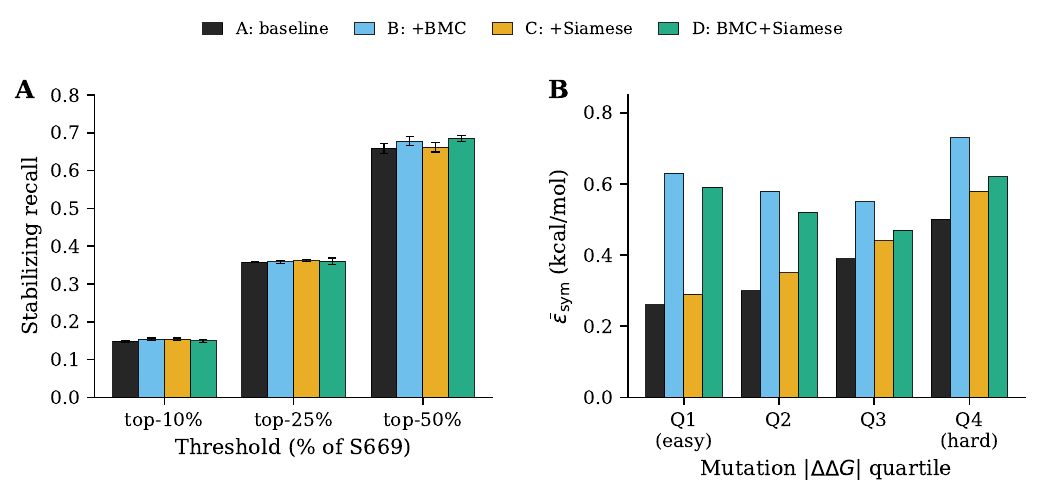}
  \caption{(\textbf{A}) Recall of truly stabilizing mutations ($\Delta\Delta G < -0.5$~kcal/mol) within the top-$k\%$ of S669
  predictions, mean $\pm$ std across three seeds. The combined configuration D improves top-50\% recall by $+0.026$; differences at top-10\% and top-25\% are within seed noise. (\textbf{B})Forward-reverse violation
  $\bar{\varepsilon}_{\mathrm{sym}}$ on Ssym, stratified by ground-truth $|\Delta\Delta G|$ quartile. The cost of distribution-balanced training is concentrated in the neutral regime (Q1) and shrinks substantially on the hardest mutations (Q4).}
  \label{fig:fig2}
\end{figure*}

In applied protein engineering workflows, global correlation matrices are often less relevant than the retrieval rate of rare stabilizing variants ($\Delta\Delta G \le -0.5$ kcal/mol). We therefore evaluate the recovery of these low-frequency events on S669 (Figure~\ref{fig:fig2} A).

% \begin{table}[h]
% \centering
% \caption{Recall of truly stabilizing mutations
% ($\Delta\Delta G > +0.5$~kcal/mol) within the top-$k\%$ of model
% predictions on S669, mean $\pm$ standard deviation across three seeds.
% The combined configuration improves top-50\% recall by $+0.026$
% relative to the baseline; differences at top-10\% and top-25\% are
% within seed-level noise on the small S669 set.}
% \label{tab:stab_recall}
% \begin{tabular}{lccc}
% \toprule
% Config & Recall @ top-$10\%$ & Recall @ top-$25\%$ & Recall @ top-$50\%$ \\
% \midrule
% A: baseline    & $0.148 \pm 0.002$ & $0.358 \pm 0.002$ & $0.659 \pm 0.013$ \\
% B: + BMC       & $0.154 \pm 0.003$ & $0.359 \pm 0.004$ & $0.678 \pm 0.012$ \\
% C: + Siamese   & $0.154 \pm 0.003$ & $0.362 \pm 0.003$ & $0.662 \pm 0.013$ \\
% D: combined    & $0.150 \pm 0.004$ & $0.360 \pm 0.009$ & $\mathbf{0.685 \pm 0.008}$ \\
% \bottomrule
% \end{tabular}
% \end{table}

The combined configuration (\textbf{D}) improves the top-50\% recall of stabilizing mutations from $0.659$ to $0.685$. In a hypothetical high-throughput screening pipeline characterizing $1{,}000$ variants, this improvement enables researchers to consistently obtain a higher yield of viable, stable candidates from the same ranked pool, roughly $\sim 2$--$3$ additional true positives per round at the conventional engineering cut-off. This improvement attenuates at highly restrictive thresholds (top-$10\%$), where predictions are dominated by trivial, solvent-exposed substitutions that can be identified by sequence consensus alone. The differentiating power of the BMC-regularized landscape emerges in the broader retrieval pool, enabling it to recover complex structural stabilizations that standard regression networks typically suppress.
\subsection{Error Stratification by Thermodynamic Magnitude}
\label{sec:res_magnitude}
Finally, we stratify anti-symmetry violations on Ssym by ground-truth thermodynamic magnitude to identify precisely where optimization breaks down (Figure~\ref{fig:fig2} B).
% \begin{table}[h]
% \centering
% \caption{Forward-reverse violation $\bar{\varepsilon}_{\mathrm{sym}}$
% on Ssym, stratified by ground-truth $|\Delta\Delta G|$ quartile.
% The gap between the BMC-based configurations and the baseline shrinks
% substantially on the hardest mutations (Q4): the cost of
% distribution-balanced training is concentrated in the easy regime
% where absolute errors are small in any case. All quantities are mean
% across three seeds.}
% \label{tab:strat_mag}
% \begin{tabular}{lcccc}
% \toprule
% Config & Q1 (easy) & Q2 & Q3 & Q4 (hard) \\
% \midrule
% A: baseline    & $0.26$ & $0.30$ & $0.39$ & $0.50$ \\
% B: + BMC       & $0.63$ & $0.58$ & $0.55$ & $0.73$ \\
% C: + Siamese   & $0.29$ & $0.35$ & $0.44$ & $0.58$ \\
% D: combined    & $0.59$ & $0.52$ & $0.47$ & $0.62$ \\
% \midrule
% Gap (D $-$ A)  & $+0.33$ & $+0.22$ & $+0.08$ & $+0.12$ \\
% \bottomrule
% \end{tabular}
% \end{table}
The baseline model exhibits increasing anti-symmetry violations as mutational magnitudes increase, growing from 0.26 kcal/mol in the most neutral quartile (Q1) to 0.50 kcal/mol in the most extreme quartile (Q4). This scaling is physically consistent, as larger free energy changes involve profound local geometric rearrangements that violate linear assumptions. When comparing the BMC-driven configurations against the baseline, we observe that the penalty incurred by distribution-balanced training is heavily concentrated in the neutral regime (+0.33 kcal/mol gap in Q1). In the extreme, high-magnitude regime (Q4), this gap narrows significantly (+0.12 kcal/mol). This stratification demonstrates that the numerical cost of regularized training is safely isolated to neutral mutations, where absolute computational errors are highly tolerable in experimental pipelines, thereby preserving fidelity in the extreme mutational regimes that govern practical protein design. 

\section{Additional negative results}
\label{sec:negative}
We evaluated several additional physically motivated
interventions that did not improve OOD performance.

First, auxiliary multitask supervision using the underlying proteolysis-derived $K_{50}$ measurements
did not improve generalization beyond the primary
BMC + Siamese configuration. Retrospective analysis
showed that the published $\Delta\Delta G$ labels
are already highly correlated with the underlying
$K_{50}$ values (Pearson $r=0.987$), indicating that
the auxiliary task provided minimal additional
information.

Second, ProteinMPNN-derived structural relaxation features were evaluated through coordinate perturbation and auxiliary feature integration. Although inference-time perturbation improved Ssym-inverse performance, all variants reduced or failed to improve wild-type-based OOD benchmarks such as S669 and S461.

We also evaluated a bias-corrected anti-symmetric (BCAS) loss of our 
own design, defined as
\begin{equation}
\mathcal{L}_{\mathrm{BCAS}}
= \alpha \Bigl[\tfrac{1}{B}\sum_i (\hat{f}_{\to}^{(i)} 
   + \hat{f}_{\leftarrow}^{(i)})\Bigr]^{2}
+ \beta \, \tfrac{1}{B}\sum_i 
   (\hat{f}_{\to}^{(i)} + \hat{f}_{\leftarrow}^{(i)})^{2},
\label{eq:bcas}
\end{equation}
with $\alpha=1.0$, $\beta=0.5$. The first term penalizes the squared batch-mean of the forward-reverse sum (the systematic bias), and the second term reduces to the standard Siamese variance penalty. BCAS substantially reduces the Ssym systematic bias (to $|<0.1|$~kcal/mol, down from $0.29$-$0.40$~kcal/mol in configurations A-D), but the bias 
reduction does not improve Spearman on OOD benchmarks (BMC+BCAS gives S669 $0.518 \pm 0.004$, matching BMC+Siamese; BMC+BCAS+OOD-margin gives $0.531 \pm 0.005$, within $0.001$ of the equivalent Siamese configuration). This is direct evidence that systematic bias and OOD generalization are decoupled in this regime.

\section{Discussion}
\label{sec:discussion}
Recent foundation model-based predictors have substantially improved protein stability prediction, yet robust generalization to unseen proteins and mutation regimes remains challenging. In this work, we investigated whether physically motivated optimization strategies can improve OOD robustness without modifying the underlying architecture. Across multiple external benchmarks, the proposed framework consistently improved performance on structurally challenging datasets such as S669 and S461, while requiring no additional architectural parameters or pretrained components. These findings suggest that optimization behavior remains an important and comparatively underexplored limitation of modern multimodal stability predictors.

An important observation emerging from our experiments is that the proposed objectives do not appear to improve performance through the exact physical mechanisms they were originally designed to enforce. Balanced Mean Squared Error improved OOD performance but produced only modest gains in stabilizing-mutation recovery. Similarly, the Siamese anti-symmetric objective altered the structure of forward-reverse prediction bias without fully eliminating systematic directional offsets. The OOD-margin regularizer likewise improved
generalization only within a narrow perturbation regime, suggesting that its effect is not simply generic noise regularization.

Architecturally-constrained methods such as JanusDDG~\citep{barducci2026janusddg} demonstrate that perfect antisymmetry can be achieved by design; our loss-level framework demonstrates that, even without exact constraint enforcement, similar gains in OOD generalization can be extracted from existing multimodal backbones at minimal additional cost.
Taken together, these observations suggest that the primary benefit of the proposed framework arises through implicit regularization rather than strict thermodynamic enforcement. Each loss component perturbs the optimizer to reduce excessive reliance on dominant in-distribution statistics and improve robustness under distribution shifts. Importantly, the observed OOD gains, therefore, cannot be explained solely through exact thermodynamic consistency. Instead, our results indicate that optimization-level thermodynamic regularization may improve transferability even when physical reversibility remains only partially satisfied.

This redistribution of optimization emphasis is accompanied by a modest reduction in in-distribution Megascale performance. We interpret this trade-off as evidence of a shift in capacity away from optimization for dominant training-set statistics toward more transferable representations. From a practical perspective, this behavior is often desirable. Real protein-engineering and variant prioritization pipelines routinely encounter proteins absent from large mutational-scanning datasets and therefore benefit more from robust transferability than from maximal in-distribution fit.

Several negative results reported here further clarify the limits of physically motivated supervision in current multimodal predictors. Neither multitask supervision using per-protein $K_{50}$ measurements nor ProteinMPNN-derived structural-relaxation features produced measurable improvements in OOD performance. These findings suggest that simply adding additional physically motivated descriptors is insufficient to improve robustness if the underlying optimization dynamics remain unchanged. More broadly, they indicate that the effective utilization of learned representations may currently be a more important bottleneck than the availability of additional structural information.

The present study has several limitations. Our analysis is limited to single-point substitutions and does not address higher-order epistatic interactions. The OOD-margin formulation also relies on a single uniform perturbation scale across all feature dimensions; adaptive perturbation strategies informed by empirical feature variance may provide further improvements.

Despite these limitations, the results presented here suggest that optimization-level physical regularization remains a promising direction for multimodal protein stability prediction. Much recent progress in the field has focused primarily on larger pretrained models and increasingly complex architectural fusion strategies. Our findings demonstrate that carefully designed optimization objectives can still extract meaningful improvements from existing multimodal representations, particularly under challenging out-of-distribution conditions.

More broadly, these observations may extend beyond protein stability prediction itself. Many scientific machine-learning systems introduce soft physical constraints during optimization under the assumption that improved physical consistency will directly translate into better generalization. Our results suggest that the relationship may instead operate indirectly through changes in optimization dynamics and representation stability. Understanding this interaction more carefully may therefore be important for scientific machine learning more broadly.
\paragraph{Code Availability.}

Source code is available at:

\url{https://github.com/shiv-ram-repo/constraint-aware-ddg}

\bibliographystyle{unsrtnat}
\bibliography{reference}

\end{document}